\documentclass[10pt,twocolumn,letterpaper]{article}

\usepackage{iccv}
\usepackage{times}
\usepackage{epsfig}
\usepackage{graphicx}
\usepackage{amsmath}
\usepackage{amssymb}

\usepackage{multirow}
\usepackage{url}
\usepackage{bm}
\usepackage{color}
\usepackage{booktabs}
\usepackage{caption}
\usepackage{floatrow}
\usepackage{authblk}
\usepackage{makecell}
\usepackage[toc,page]{appendix}
\newcommand{\dan}[1]{{\color{black}#1}}
\newcommand{\tang}[1]{{\color{black}#1}}

\usepackage{hyperref}
\hypersetup{
    colorlinks=true,
    linkcolor=blue,
    filecolor=magenta,      
    urlcolor=magenta}

\iccvfinalcopy 

\def\iccvPaperID{5406} 

\ificcvfinal\fi

\begin{document}

\title{SA-ConvONet: Sign-Agnostic Optimization of \\ Convolutional Occupancy Networks}
\author[1,4]{Jiapeng Tang} 
\author[1]{Jiabao Lei}
\author[2]{Dan Xu} 
\author[4]{Feiying Ma}
\author[1,5,6]{Kui Jia}
\author[3,4]{Lei Zhang}
\affil[1]{School of Electronic and Information Engineering, South China University of Technology}
\affil[2]{Department of Computer Science and Engineering, HKUST, HK}
\affil[3]{Department of Computing, The Hong Kong Polytechnic University, HK}
\affil[4]{DAMO Academy, Alibaba Group}
\affil[5]{Pazhou Lab, Guangzhou, China} 
\affil[6]{Peng Cheng Laboratory, Shenzhen, China}


\twocolumn[{%
\renewcommand\twocolumn[1][]{#1}%
    \maketitle
    
    \begin{center}
        \vspace{-40pt}
        \centering
        \includegraphics[scale=0.15]{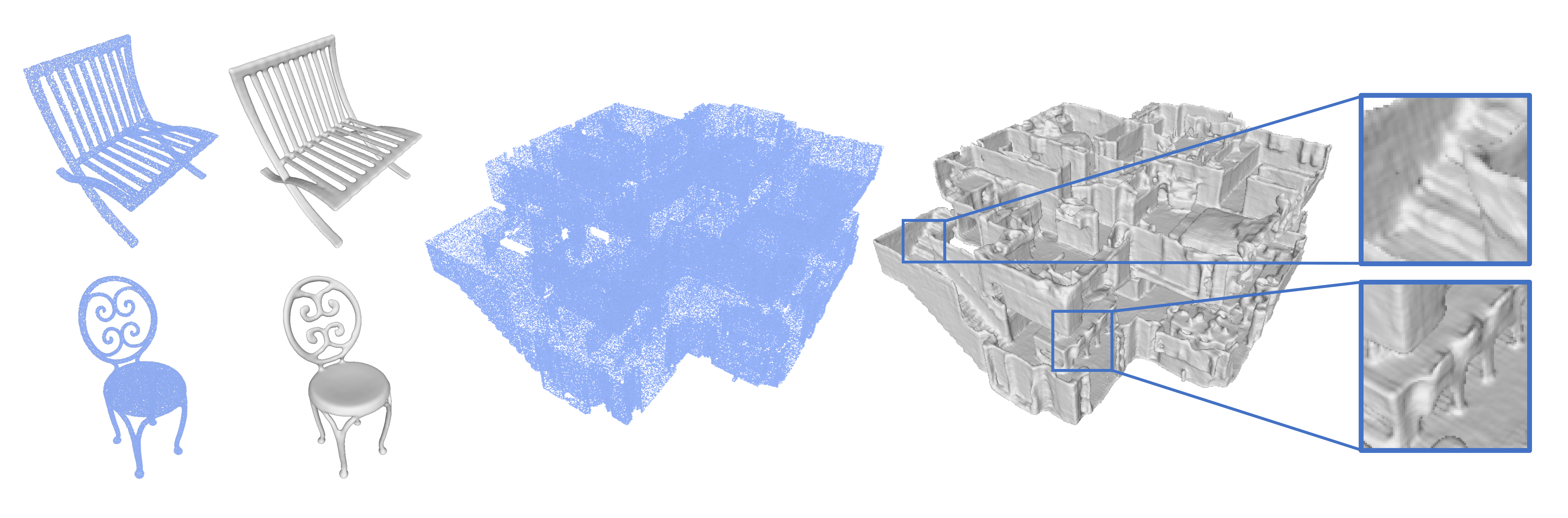}
        \vspace{-10pt}
        \captionof{figure}{Given an un-oriented point cloud of a complex object (left) or a large-scale scene (right), our method can reconstruct \tang{an} accurate surface mesh \emph{without the use of oriented normals}.}
        \label{fig:teaser}
    \end{center}
}]

\renewcommand{\thefootnote}{\fnsymbol{footnote}}
\footnotetext{Correspondence to Dan Xu and Kui Jia.}

\ificcvfinal\thispagestyle{empty}\fi


\begin{abstract}
    
    Surface reconstruction from point clouds is a fundamental problem in the computer vision and graphics community. Recent state-of-the-arts solve this problem by individually optimizing each local implicit field during inference. Without considering the geometric relationships between local fields, they typically require accurate normals to avoid the sign conflict problem in overlapped regions of local fields, which severely limits their applicability to raw scans where surface normals could be unavailable. 
    Although SAL breaks this limitation via sign-agnostic learning, further works still need to explore how to extend this technique for local shape modeling.
    To this end, we propose to learn implicit surface reconstruction by sign-agnostic optimization of convolutional occupancy networks, to simultaneously achieve advanced scalability to large-scale scenes, generality to novel shapes, and applicability to raw scans in a unified framework. Concretely, we achieve this goal by a simple yet effective design, which further optimizes the pre-trained occupancy prediction networks with an unsigned cross-entropy loss during inference. The learning of occupancy fields is conditioned on convolutional features from an hourglass network architecture.
    Extensive experimental comparisons with previous state-of-the-arts on both object-level and scene-level datasets demonstrate the superior accuracy of our approach for surface reconstruction from un-orientated point clouds. The code is available at {\url{ https://github.com/tangjiapeng/SA-ConvONet}}.

\end{abstract}

\section{Introduction}
\label{SecIntro}
    Surface reconstruction from point clouds is of significance to perceive and understand surrounding 3D worlds for intelligent systems, which plays a \dan{fundamental} role in numerous practical applications, such as computer-aided design, 3D printing, and robotics grasping. Recently, this problem has attracted \dan{wide} attention as inexpensive and portable commodity scanners such as the Microsoft Kinect make it much easier to acquire 3D point clouds. Classical methods~\cite{alexa2003computing, carr2001reconstruction, levin2004mesh, kazhdan2006poisson, kazhdan2013screened} \dan{tackle} this problem by mathematical optimization according to pre-defined geometric priors, while learning-based methods ~\cite{groueix2018atlasnet, chibane2020implicit, mescheder2019occupancy, peng2020convolutional} choose to learn geometric priors from large-scale 3D datasets in a data-driven manner. 
    Recently, \dan{representing 3D surface as an implicit field} has gained \dan{large} popularity~\cite{chen2019learning, mescheder2019occupancy, park2019deepsdf, saito2019pifu, xu2019disn, chibane2020implicit, jiang2020local, chabra2020deep, erler2020points2surf, tretschk2020patchnets, peng2020convolutional, tang2021skeletonnet}. Compared to other shape representations \dan{such as} voxel~\cite{choy20163d, wu2016learning}, octree~\cite{riegler2017octnet, tatarchenko2017octree, wang2018adaptive, hane2017hierarchical}, point cloud~\cite{fan2017point} and mesh~\cite{groueix2018atlasnet, wang2018pixel2mesh, kato2018neural, tang2019skeleton, pan2019deep, tang2021skeletonnet}, continuous implicit fields can enable surface reconstruction with infinite resolution and arbitrary topology. 
    
    A lot of methods have been proposed to \dan{advance} the development of implicit surface reconstruction from various respects in terms of improving scalability, generality, and applicability. \dan{However,} there is still not an approach \dan{in the literature} to simultaneously achieve all these \dan{objectives with satisfactory performance}.
    Targeting better scalability \tang{to} large-scale scenes, \dan{several} approaches~\cite{jiang2020local, chabra2020deep,  tretschk2020patchnets, peng2020convolutional, erler2020points2surf} learn local implicit fields and model a global shape as a composition of local surface geometries, rather than conducting global shape reasoning from a latent code.
    \dan{Towards} better generality \tang{to} novel shapes, \dan{some works}~\cite{park2019deepsdf, gropp2020implicit,jiang2020local, chabra2020deep, tretschk2020patchnets, yang2020deep} attempt to optimize the pre-trained priors at test time to \tang{obtain a better} solution for each given input, instead of strictly respecting the learned priors. 
    Existing state-of-the-art methods~\cite{jiang2020local, chabra2020deep, tretschk2020patchnets} improve both scalability and generality \dan{via individual optimization of } each local implicit field during inference.
    However, without explicitly considering geometric relationships between local fields,
    they heavily rely on accurate normals to avoid the sign conflict problems in the overlapped regions of local fields.  
    Although SAL~\cite{atzmon2020sal} breaks this limitation via sign-agnostic learning that improves the applicability to real-world scans where surface normals are unavailable, it can only perform global shape modeling. \tang{Further works still need to explore how to extend this technique for local shape modeling.}
    
    To this end, we propose to learn implicit surface reconstructions by sign-agnostic optimization of convolutional occupancy networks~\cite{peng2020convolutional}, to simultaneously achieve \dan{the three important reconstruction objectives,} \ie~advanced generality, specialty, and applicability in a unified framework. \dan{We achieve} this goal by a simple yet effective solution that further optimizes the pre-trained occupancy prediction networks via sign-agnostic learning. The learning of occupancy fields is conditioned on convolutional features from an hourglass network (\eg~U-Net~\cite{ronneberger2015u}).
    Our solution is motivated by two key characteristics.
    The first characteristic is that, after being pre-trained on the accessible datasets with ground-truth signed fields, the occupancy decoder can provide a signed field as initialization for the test-time optimization. Thus we can further apply unsigned objectives to optimize occupancy prediction networks, maximizing the consistency between the desired iso-surface with the observed un-oriented point cloud.
    The second characteristic is that, the U-Net~\cite{ronneberger2015u} aggregates both local and global information in an hourglass convolutional manner. The use of local shape features not only preserves the fine-grained geometries, but also enables the surface recovery of large-scale indoor scenes.
    The integrated global shape features can enforce geometric consistency between learned local geometries and guarantee the assembly of local fields as a globally consistent one, \dan{although} we do not \dan{utilize} guidance from additional normal information.
    As shown in Figure~\ref{fig:teaser}, we can reconstruct surfaces with \dan{fine} details directly from un-oriented point clouds \emph{without the use of normals}, \dan{for both} complicated objects to large-scale scenes.
    
    Extensive \dan{experimental} comparisons \dan{with} state-of-the-arts on both object-level and scene-level datasets, including ShapeNet~\cite{chang2015shapenet}, synthetic indoor scene dataset~\cite{peng2020convolutional}, and real-world scene datasets (ScanNet~\cite{dai2017scannet} and Matterport3D~\cite{chang2017matterport3d}) demonstrate the superior \dan{performance} of our approach for surface reconstruction from un-oriented point clouds.

    
\section{Related Work}
In this section, we briefly review existing methods for surface reconstruction from raw point clouds. Specifically, we only review those implicit reconstruction methods that find a field function (occupancy function or signed distance function) to approximate the given point cloud.
    
\noindent \textbf{Classic Optimization-based Surface Reconstruction}
 Computing a continuous surface from its discrete approximation is a severely ill-posed problem, since there could be infinitely possible solutions. Classical methods ~\cite{alexa2003computing, carr2001reconstruction, levin2004mesh, kazhdan2006poisson, kazhdan2013screened} formulate this task as a mathematical optimization problem and try to solve this problem \dan{utilizing} pre-defined geometric priors such as local linearity and smoothness. 
 There have been a number of representative reconstruction methods \dan{such as} Radius Basis Function (RBF)~\cite{carr2001reconstruction}, Moving Least Square (MLS)~\cite{alexa2003computing}, and Poisson Surface Reconstruction (PSR)~\cite{kazhdan2006poisson, kazhdan2013screened}.
 The RBF~\cite{carr2001reconstruction} represents surface as \dan{a} linear combination of a series of radial basis functions; \dan{the} MLS~\cite{alexa2003computing} fits observed points via finding those constituent spatially-varying polynomials; the PSR~\cite{kazhdan2006poisson, kazhdan2013screened} models the surface reconstruction as \dan{a} Poisson's equation.
  

\noindent \textbf{Learning-based Surface Reconstruction}
More recently, driven by large-scale 3D datasets \dan{(e.g.  ShapeNet~\cite{chang2015shapenet})}, neural networks have achieved \dan{notable} successes in the field of implicit surface reconstruction from point clouds, \dan{ranging from} global to local field modeling. The global models~\cite{mescheder2019occupancy, chen2019learning} intrinsically perform shape retrieval in the latent space~\cite{tatarchenko2019single}, leading to limited generality to represent unseen shapes and restricted power to capture complex details. These drawbacks can be \dan{resolved} by the local models~\cite{chibane2020implicit, saito2019pifu, xu2019disn, peng2020convolutional, erler2020points2surf} that focus on local geometry modeling. Our method also adopts the manner of the local implicit field learning.  Thus it possesses the capability of representing large-scale scenes. Another advantage is the better generality to unseen shapes, which can provide relatively good \dan{initialization} of signed fields for the test-time optimization.

\noindent \textbf{Combination of \dan{Data-Driven} Priors and Optimization}
The above-mentioned learning methods~\cite{chibane2020implicit, saito2019pifu, xu2019disn, peng2020convolutional} fix the learned priors during inference. \dan{Specifically}, they directly obtain a 3D surface via a single feed-forward pass.
As the pre-trained priors are fixed, they have difficulty \dan{in} generalizing well to unseen shapes that are dissimilar to the training samples. 
Some existing approaches ~\cite{park2019deepsdf, jiang2020local, chabra2020deep, tretschk2020patchnets, xu2019geometry, yang2020deep, tang2021learning} try to combine the data-driven priors with optimization strategy at the test phase to acquire better results for each given input.  
Among them, the methods of local field optimization, including LIG~\cite{jiang2020local}, DeepLocalSDF~\cite{chabra2020deep}, and PatchNet~\cite{tretschk2020patchnets}, can achieve state-of-the-art performance. However, they require \emph{additional surface normals} to solve the sign flipping problem when assembling local fields into a globally consistent one, which seriously limits their applicability to raw scans that lack reliable and accurate surface normals. Our method also belongs to this line of \dan{research}. However, \dan{in contrast to} them, we optimize local implicit fields that are conditioned on the convolutional features learned in an hourglass manner. Since the global consistency between local fields can be effectively maintained in the process of hourglass convolutional feature learning, we can always guarantee the local field \dan{assemblies} as a globally consistent one during optimization, although we do not have the guidance of surface normals.

\noindent \textbf{Sign-Agnostic Surface Reconstruction}
The raw point clouds scanned by sensing devices usually lack oriented normals. Although we can approximate them via normal estimation methods~\cite{hoppe1992surface, guennebaud2007algebraic,Sheng2019Unsupervised, guerrero2018pcpnet, ben2020deepfit}, the normal estimation errors can cause degenerated surfaces. Thus it is more appealing to model surface directly from un-oriented points in a sign-agnostic manner~\cite{atzmon2020sal, atzmon2020sal++}. The SAL~\cite{atzmon2020sal} avoids the use of surface normals by properly initializing the implicit decoder network, such that they can produce signed solutions of implicit functions only using unsigned objectives. 
Our key idea of sign-agnostic implicit field optimization is similar to SAL. With the assistance of auxiliary datasets with ground-truth signed implicit fields, the occupancy decoder can be trained to represent signed fields. Given signed implicit fields as \dan{initialization} for the test-time optimization, we further adapt the pre-trained priors to the given input, by applying the unsigned cross-entropy loss to align the desired iso-surface with the observed un-oriented point cloud. 
\tang{A concurrent work of SAIL-S3~\cite{zhao2021sign} extended the geometric initialization of SAL~\cite{atzmon2020sal} for local signed field learning from un-oriented point clouds. However, it still requires a post-optimization stage to avoid the local sign flipping issue.}


\section{Approach}
\label{SecApp}

\subsection{Overview}
\label{SubSecOver}
    Given a set of observed points $\mathcal{P} = \{\mathbf{p}_i \in \mathbb{R}^3 \}_{i=1}^n$, the goal of our method is to reconstruct a surface $\mathcal{S}$ that is as similar as possible to the underlying surface $\hat{\mathcal{S}}$. We choose to approximate the signed implicit field representation $\hat{\mathbf{O}}$ of $\hat{\mathcal{S}}$ by predicting a neural implicit field $\mathcal{O}$ due to its advantage of reconstructing surfaces with infinite resolution and unrestricted topology.
    Our goal is to simultaneously achieve advancements in all three respects, \tang{\ie scalability to large-scale scenes, generality to novel shapes, and applicability to raw scans}. 
    Towards this goal,  we propose a simple yet effective solution of learning implicit surface reconstructions by sign-agnostic optimization of convolutional occupancy networks.  
    
\begin{figure*}[t]
    \centering
    \includegraphics[scale=0.45]{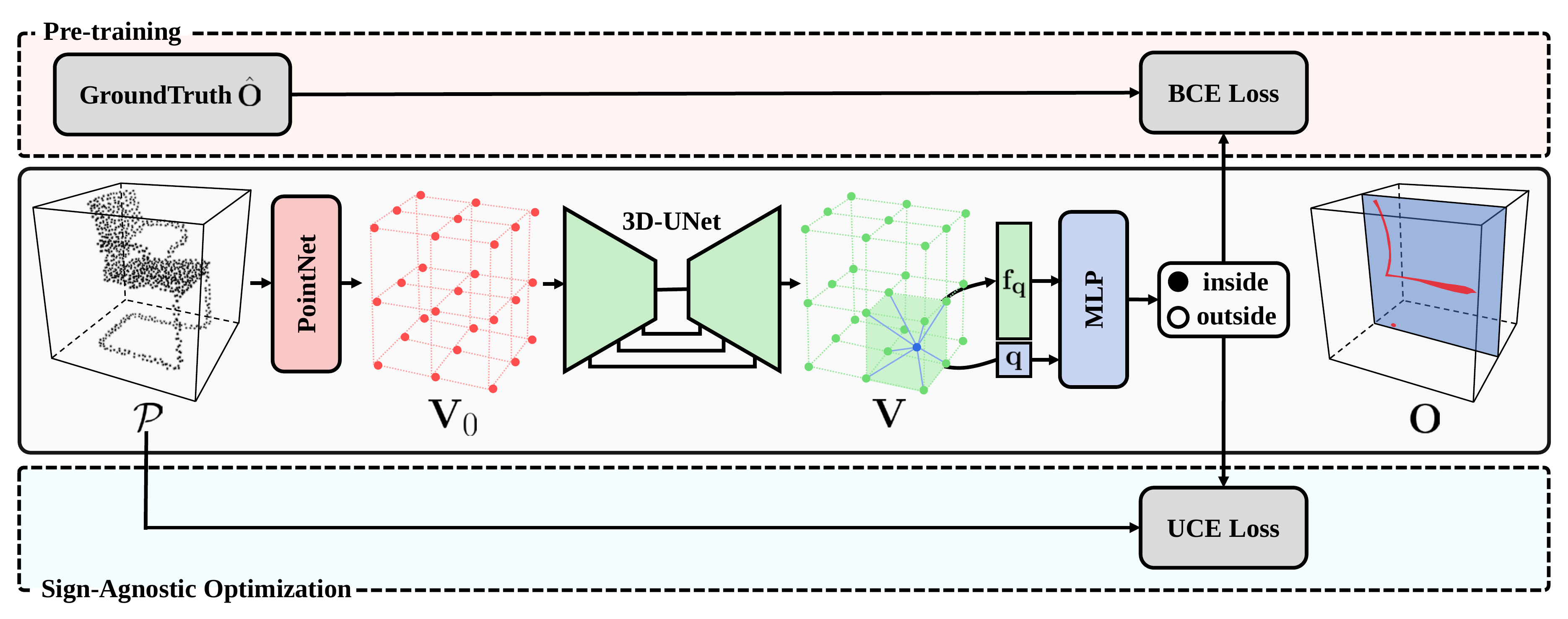}
    \vspace{0.1cm}
    \caption{\textbf{Method Overview.}   
    Our approach is built \dan{upon} the convolutional occupancy networks (CONet)~\cite{peng2020convolutional} (middle) that predicts an occupancy field $\mathbf{O}$ based on convolutional features $\mathbf{V}$ extracted from the input point cloud $\mathcal{P}$ via a cascaded network of PointNet and 3D U-Net.
    We \dan{first} pre-train the CONet~\cite{peng2020convolutional} on the accessible datasets with ground-truth $\hat{\mathbf{O}}$ using the standard binary cross-entropy (BCE) loss (top). During inference, the proposed sign-agnostic optimization further \dan{fine-tunes} the whole network parameters via unsigned cross-entropy (UCE) loss to improve the accuracy of $\mathbf{O}$ (bottom). }
    
    \label{fig:pipeline}
    \vspace{-0.5cm}
\end{figure*}

    The overall pipeline \dan{of the proposed approach} is shown in Figure~\ref{fig:pipeline}.
    Our approach consists of two stages, \dan{namely} convolutional occupancy field pretraining and designed sign-agnostic, test-time implicit surface optimization.
    The former \dan{stage} is responsible \dan{for learning} the local shape priors with global consistency \dan{constraints}, and \dan{provides} relatively reasonable signed fields as \dan{initialization} for the latter \dan{stage}, \dan{which} further optimizes the whole network using the unsigned cross-entropy loss to improve the accuracy of $\mathbf{O}$. 
    \dan{We present the details of the proposed approach} in the following two sections, \ie Section~\ref{SecConvONet} and Section~\ref{SecSurfFit}.

\subsection{Convolutional Occupancy Fields Pre-training}
\label{SecConvONet} 
    
    \subsubsection{Convolutional Feature Learning}
    \label{SecUNet} 
    As shown in Figure~\ref{fig:pipeline}, we \dan{first} process the given input $\mathcal{P}$ by a shallow PointNet~\cite{charles2017pointnet} to obtain point-wise features. Then, we convert them to volumetric features with \dan{a dimension} of $H \times W \times D$\dan{,} by encapsulating local neighborhood information within a cell. \dan{Specifically,} we integrate all point features belonging to the same voxel cell using \dan{the} average pooling.
    To integrate both global and local information, we use a 3D-UNet to process $\mathbf{V}_0$ to obtain $\mathbf{V}$. Due to the \dan{issue of memory overhead} of 3D-CNN, we set $H=W=D=64$\dan{, and} the depth of the 3D U-Net is set to 4 such that the receptive field is equal to the size of $\mathbf{V}_0$.  
    \dan{Due to} the \dan{translation} equivariance of convolution operations and rich shape features integrated by the hourglass \dan{netowork architecture,} U-Net, we can enable scalable surface reconstruction for large-scale scenes. 
    
    \subsubsection{Occupancy Field Predicting}
    \label{SecOccPred} 
    Based on the obtained volumetric features $\mathbf{V}$, we can predict the occupancy probability $\mathbf{O}({\mathbf{q}}) \in \mathbb{R}^3$ of a point $\mathbf{q}$ randomly sampled in 3D space. To do this, we \dan{first} perform trilinear interpolation to query the feature vector $\mathbf{f_q}$ from $\mathbf{V}$ according to the coordinate of $\mathbf{q}$, and then feed $\mathbf{q}$ and $\mathbf{f_q}$ into the occupancy decoder $g$ that is implemented as a light-weight network of multi-layer perceptron (MLP):
    \begin{equation}
    \label{EqnOccPred}
        \mathbf{O}({\mathbf{q}})  = \mathrm{sigmoid}\left( g(\mathbf{q}, \mathbf{f_q}) \right) \in (0, 1),
    \end{equation}
    where the occupancy probability of $\mathbf{q}$ is the $\mathrm{sigmoid}$ activation of final output logit $g(\mathbf{q}, \mathbf{f_q})$.
    
    \subsubsection{Loss Function}
    \label{SecLoss} 
    During training, we uniformly sample some points $\mathcal{Q}$ within the bounding volume of watertight mesh and compute their ground truth of occupancy values. And we punish the discrepancy between the predicted and \dan{the} true occupancy values by \dan{a} loss function \dan{written as}:
    
    \begin{equation}
    \label{Eqnbce}
        \mathcal{L}(\mathbf{O}, \hat{\mathbf{O}})  = {\sum_{\mathbf{q} \in \mathcal{Q}}} 
            \texttt{BCE}\left(\mathbf{O}({\mathbf{q}}), \hat{\mathbf{O}}({\mathbf{q}})\right),
    \end{equation}
    where $\texttt{BCE}(x, y) = -y \mathrm{log}x - (1-y) \mathrm{log}(1-x)$ denotes the standard binary cross-entropy.
    
 \subsection{Sign-Agnostic Implicit Surface Optimization}
 \label{SecSurfFit}
    In the inference stage, we can directly \dan{produce} the implicit field through a single \dan{feed-forward pass}. But we may not get satisfactory results if the given inputs are out of pre-trained priors.
    In order to improve the generality to unseen shapes, one can further optimize the pre-trained model for the given input. 
    But we cannot apply the loss function in Equation \ref{Eqnbce} to supervise the network finetuning, because surface normals associated with \dan{observed} points are not available, which causes the unavailability of in-out fields. Although we can choose to estimate the normals from $\mathcal{P}$, the normal estimation errors would increase the difficulty of recovering clean surfaces.

    However, the requirements of normals can be avoided by sign-agnostic optimization of occupancy field learning from hourglass convolutional networks.
    According to SAL~\cite{atzmon2020sal}, we know that by properly initializing network parameters, the implicit decoder can represent the signed field of a unit sphere, which helps us obtain signed solutions by unsigned learning objectives.
    Similarly, the pre-trained occupancy decoder can produce signed fields as \dan{initialization} for the test-time optimization. As such, we can directly employ the unsigned cross-entropy loss to \dan{obtain consistency constraints} between the occupancy field and unsigned inputs\dan{,} without the use of surface normals. Besides, global consistency \dan{among} local geometries can \dan{always be} enforced during the optimization stage, because the \dan{features} from $\mathbf{V}$ are decoded from the same global features. Thus, without the guidance of normals,  \tang{we can still guarantee globally consistent local field assemblies}.
    \dan{Specifically}, the unsigned cross-entropy (UCE) loss \dan{is} formulated as:
    \begin{equation}
     \label{EqnUsgBce}
        \mathcal{L}_{uce} = {\sum_{\mathbf{q} \in \mathcal{Q}}} 
            \texttt{BCE}\left(\mathbf{O}^{\dag}({\mathbf{q}}), \hat{\mathbf{O}}^{\dag}({\mathbf{q}})\right),
    \end{equation} 
    where prediction $\mathbf{O}^{\dag}({\mathbf{q}})$ and target $\hat{\mathbf{O}}^{\dag}({\mathbf{q}})$ are given by
    \begin{equation}
    \label{EqnAbsOcc}
        \mathbf{O}^{\dag}({\mathbf{q}}) = \mathrm{sigmoid} \big( | g(\mathbf{q}, \mathbf{f_q}) | \big) \in [0.5, 1),
    \end{equation}
    \begin{equation}
    \label{EqnUsgGT}
        \hat{\mathbf{O}}^{\dag}({\mathbf{q}}) = 
                        \left\{
                            \begin{aligned}
                                 0.5, \hspace{5pt} & \textrm{for} \ \mathbf{q} \in \mathcal{Q}_{\hat{\mathcal{S}}} \\
                                 1.0, \hspace{5pt} & \textrm{for} \ \mathbf{q}\in \mathcal{Q}_{\backslash {\hat{\mathcal{S}}}}
                            \end{aligned}
                        \right.
        ,
    \end{equation}
    where $\mathcal{Q}_{\hat{\mathcal{S}}}$ is \dan{a} point set \dan{obtained} from \dan{the} ground-truth surface $\hat{\mathcal{S}}$, and $\mathcal{Q}_{\backslash {\hat{\mathcal{S}}}}$ is \dan{a} point set sampled from non-surface volume $\mathcal{Q} \setminus \hat{\mathcal{S}}$. As $\hat{\mathcal{S}}$ is unknown at the test phase, we consider the observed surface $\mathcal{P}$ as an approximation of $\hat{\mathcal{S}}$, and identify randomly sampled points in 3D space as  non-surface points $\mathcal{Q}_{\backslash \mathcal{S}}$.
    More specifically, we  force the observed surface $\mathcal{P}$ to align with the 0.5 level set of occupancy field, and the signed occupancy values of non-surface points to be either 0 or 1.
    
    After the \dan{sign-agnostic optimization of the} implicit field, we apply the Multiresolution IsoSurface Extraction (MISE)~\cite{mescheder2019occupancy} and marching cubes~\cite{lorensen1987marching} to extract surface meshes as \dan{the} final reconstruction results. 

\section{Experiments}
\label{SecExp}

\noindent \textbf{Datasets} We validate the \dan{efficiency} of our method on experiments of both object-level and scene-level surface reconstruction \dan{tasks}. For the former \dan{task}, we conduct \dan{comparison} on the \emph{chair} category of \dan{the} ShapeNet~\cite{chang2015shapenet} dataset. 
For the latter \dan{task}, we use the synthetic indoor scene dataset ~\cite{peng2020convolutional}.
The split of train/val/test sets \tang{follows} the same setting in  CONet~\cite{peng2020convolutional}. For each dataset, we randomly select 50 models from the test set to conduct quantitative evaluations.
 We use point clouds of size 30, 000 sampled from true surfaces as inputs. Finally, we compare the synthetic-to-real generality by conducting experiments on ScanNet-V2~\cite{dai2017scannet} and Matterport3D~\cite{chang2017matterport3d} datasets.

\noindent \textbf{Implementation Details}  
We \dan{first} pre-train the convolutional occupancy networks \dan{with} a batch size of 32 and \dan{a} learning rate of $1 \times 10^{-4}$ for \dan{overall} 300k iterations.
During the sign-agnostic optimization, the whole network is further optimized by the objective \dan{described} in Equation~\ref{EqnUsgBce} using a batch size of 16 for 1000 iterations. The initial learning rate is set to $3 \times 10^{-5}$, and decays by 0.3 every 400 iterations.
We set $|\mathcal{Q}_{\hat{\mathcal{S}}}| = 512$ and $|\mathcal{Q}_{\backslash {\hat{\mathcal{S}}}}| = 1,536$ in Equation~\ref{EqnUsgGT}.

\noindent \textbf{Baselines}
We conduct \dan{comparison} with three categories of existing methods, \dan{i.e.} classic optimization-based methods such as Screened Poisson Surface Reconstruction (SPSR)~\cite{kazhdan2013screened},
deep optimization-based methods such as Sign-Agnostic Learning (SAL)~\cite{atzmon2020sal} and Implicit Geometric Regularization (IGR)~\cite{gropp2020implicit},
learning-based methods such as Occupancy Networks (ONet)~\cite{mescheder2019occupancy} and Convolutional Occupancy Networks (CONet)~\cite{peng2020convolutional}, 
\dan{and} methods \dan{focusing on} optimizing data-driven \dan{priors} such as Local Implicit Grid (LIG)~\cite{jiang2020local}. For SAL and IGR, we directly fit the neural implicit field to the observed point cloud. For ONet, CONet, and LIG, the evaluations are based on their provided pre-trained models.
\dan{Specifically,} we summarize their respective characteristics in Table \ref{table:compare_chara}. For methods that require oriented surface normals, we follow~\cite{hoppe1992surface} to estimate un-oriented normals and then reorient their directions.

    \begin{table}
        \centering
        \scalebox{0.9}{
        \begin{tabular}{l|c c c}
        \toprule
        \makecell{Methods \\} & \makecell{Without \\ normals} & \makecell{Optimization of \\ network parameters} & \makecell{Local geometry \\ modeling} \\
        \hline \hline
        SPSR~\cite{kazhdan2013screened} & $\times$  & $\checkmark$ & $\checkmark$  \\
        \hline
        ONet~\cite{mescheder2019occupancy} & $\checkmark$ &  $\times$  & $\times$ \\
        SAL~\cite{atzmon2020sal} & $\checkmark$ & $\times$  &  $\times$ \\
        IGR~\cite{gropp2020implicit} & $\checkmark$  & $\checkmark$   & $\times$ \\
        \hline
        CONet~\cite{peng2020convolutional} & $\checkmark$ &  $\times$  & $\checkmark$ \\
        LIG \cite{jiang2020local} &  $\times$ &  $\checkmark$  & $\checkmark$ \\
        \hline
        Ours & $\checkmark$ & $\checkmark$  & $\checkmark$ \\
        \bottomrule
        \end{tabular}
        }
        \vspace{0.1cm}
        \caption{
        \tang{Working condition summary of different methods.
        Note that our method is the first to maximize the scalability to large-scale scenes, generality to novel shapes, and applicability to real-world scans in a unified framework by performing local geometry modeling, optimizing network parameters while not requiring normals during inference.}
        }
        \label{table:compare_chara}
        \vspace{-0.3cm}
    \end{table}

\noindent \textbf{Evaluation Metrics}
We consider Chamfer Distance (CD $\times 0.01$), Normal Consistency (NC $\times 0.01$), and F-score (FS $\times 0.01$) as primary evaluation metrics. The F-score is reported with thresholds of $\tau$ and $2\tau$ ($\tau = 0.01$).
\dan{The} quantitative results between two point clouds are measured from randomly sampled ten thousand surface points. For the CD, the lower is better. For NC and FS, the higher is better.

\section{Object-level Reconstruction}
\label{SubSecExpObj}
      
\begin{table}[htbp]
    \vspace{-0.3cm}
        \centering
            \begin{tabular}{l|c c c c}
            \toprule
            Methods  & CD $\downarrow$ & NC $\uparrow $ & FS ($\tau$) $\uparrow$ &  FS (2$\tau$) $\uparrow$ \\ 
            \hline \hline
            SPSR \cite{kazhdan2013screened} & 1.923 & 81.54  & 80.86  & 85.13 \\     
            \hline
            ONet\cite{mescheder2019occupancy} & 1.117  & 84.58  & 62.35  & 86.57 \\  
            SAL \cite{atzmon2020sal}          & 2.418  & 78.67  & 54.33  & 73.70 \\  
            IGR \cite{gropp2020implicit}   & 2.678  & 75.97  & 69.02  & 76.01 \\ 
            \hline
            CONet \cite{peng2020convolutional}  & 0.821  & 91.12  & 74.73 & 96.85 \\ 
            LIG \cite{jiang2020local}      & 2.200  & 80.35   & 60.62   & 65.99 \\ 
            \hline
            Ours & \textbf{0.522} & \textbf{93.51} & \textbf{97.16} & \textbf{99.37} \\  
            \bottomrule
            \end{tabular}
        \vspace{0.1cm}
        \caption{
        Quantitative \dan{comparison} for surface reconstruction from un-oriented point clouds on the ShapeNet-chair. }
        \label{table:chair}
    \vspace{-0.3cm}
\end{table}

    \begin{figure*}[!htb]
            \centering 
            \includegraphics[scale=0.135]{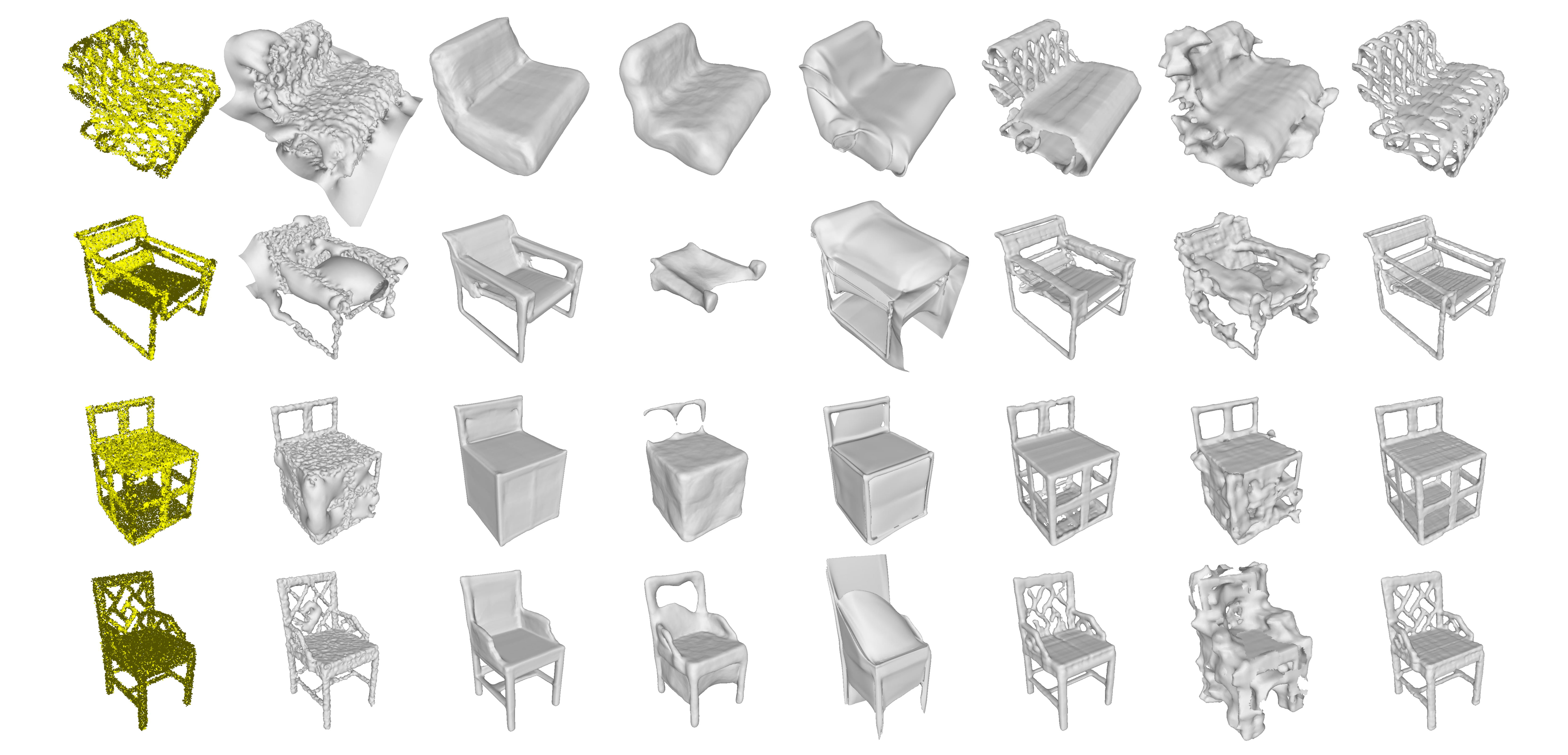}
                \begin{tabular}{p{60pt}p{53pt}p{53pt}p{53pt}p{53pt}p{53pt}p{53pt}p{53pt}}
                 \quad \quad Input PC  & 
                 \quad SPSR \cite{kazhdan2013screened} &
                 \quad ONet \cite{mescheder2019occupancy} & 
                 \quad SAL \cite{atzmon2020sal}  & 
                 IGR \cite{gropp2020implicit} & 
                 CONet \cite{peng2020convolutional}  &
                 LIG \cite{jiang2020local} &  
                 Ours
                \end{tabular}
                \caption{\textbf{Object-level Reconstruction on ShapeNet.} Qualitative \dan{comparison} for surface reconstruction from un-orientated point clouds of ShapeNet-chair~\cite{dai2017scannet}.}
                \label{fig:chair}
            
    \end{figure*}

    \begin{figure*}[!htb]
            \centering 
                \includegraphics[scale=0.127]{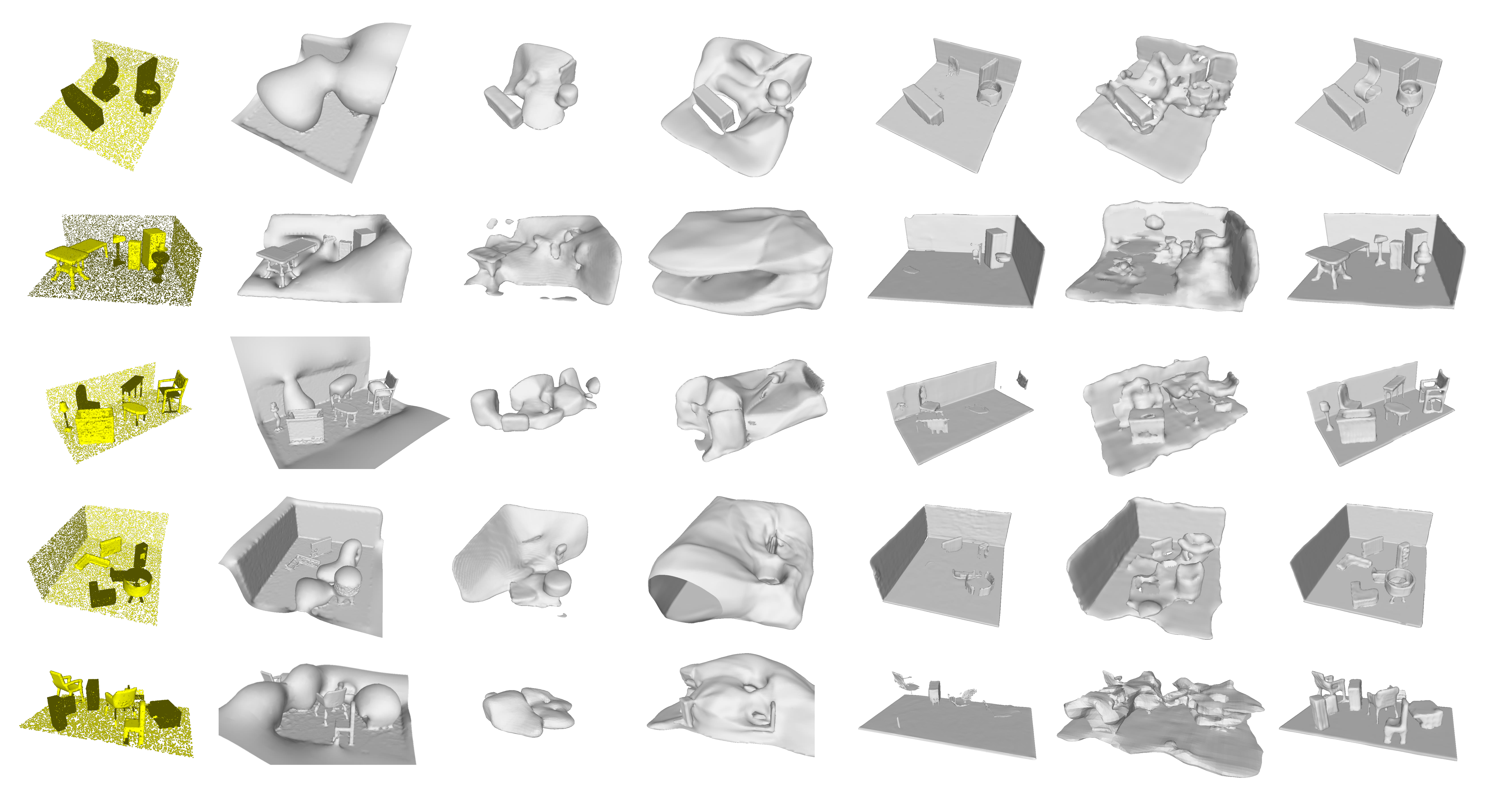}
                \begin{tabular}{p{62pt}p{62pt}p{62pt}p{62pt}p{62pt}p{62pt}p{62pt}}
                     \quad Input PC & 
                     \quad SPSR \cite{kazhdan2013screened} &
                     \quad SAL \cite{atzmon2020sal}  & 
                     IGR \cite{gropp2020implicit} & 
                     CONet \cite{peng2020convolutional}  &
                     LIG \cite{jiang2020local} &  
                     Ours
                    \end{tabular}
                \caption{\textbf{Scene-level Reconstruction on synthetic rooms.} Qualitative \dan{comparison} for surface reconstruction from un-orientated point \dan{clouds} on the synthetic room indoor scene dataset provided by~\cite{peng2020convolutional}.}
                \label{fig:synroom}
    \end{figure*}
    
    We \dan{first} conduct the object-level reconstruction experiments. 
    To \dan{simulate} the influence of sensing noises during real scans acquisition, we perturb 
    the input by gaussian noise with zero mean and standard deviation 0.05.
    
    As shown in Figure~\ref{fig:chair},  our approach has demonstrated superiority in terms of visual quality in the complex object reconstruction.
    Compared to the methods of only using global shape features such as ONet~\cite{mescheder2019occupancy}, SAL~\cite{atzmon2020sal}, and IGR~\cite{gropp2020implicit}), ours \dan{is} more capable of recovering complicated geometries because we utilize rich shape features, including \dan{both} local \dan{and} global information. Besides, instead of strictly respecting the learned \dan{priors} like CONet~\cite{peng2020convolutional}, more faithful surface details (e.g. slender bars and tiny holes) can be preserved by breaking the barrier of pre-trained \dan{priors} during inference.
    In addition, SPSR~\cite{kazhdan2013screened}, IGR~\cite{gropp2020implicit}, and LIG~\cite{jiang2020local} tend to produce degenerated meshes, caused by inaccurate normal orientation estimation.  However, it has no negative effects on our results, as we do not use the additional information during the test-time optimization.
    Our superiority is also verified by numerical results reported in Table~\ref{table:chair}, where our method outperforms existing state-of-the-art methods by large margins.

\section{Scene-Level Reconstruction}
    \begin{table}[htbp]
    \vspace{-0.3cm}
    \centering
        \begin{tabular}{l|cccc}
        \toprule
        Methods  & CD $\downarrow$ & NC $\uparrow $ & FS ($\tau$) $\uparrow$ & FS (2$\tau$) $\uparrow$  \\
        \hline\hline
        SPSR \cite{kazhdan2013screened} & 2.083 & 78.21  & 76.17 & 81.22  \\
        \hline
        SAL \cite{atzmon2020sal}          & 2.720  & 73.85 & 40.47  &  59.79   \\
        IGR \cite{gropp2020implicit}   & 1.923 & 77.94 & 74.02 & 81.23  \\
        \hline
        CONet \cite{peng2020convolutional} & 2.020 & 83.43 & 73.28 &  81.74  \\
        LIG \cite{jiang2020local}      & 1.953 & 79.82 & 62.46 &  70.96  \\
        \hline
        Ours & \textbf{0.495} & \textbf{90.04} &  \textbf{93.85} & \textbf{98.82}   \\
        \bottomrule
        \end{tabular}
    \vspace{0.1cm}
    \caption{
    Quantitative results for surface reconstruction from un-oriented point clouds on the synthetic room dataset \cite{peng2020convolutional}.} 
    \label{table:synthetic_room}
    \vspace{-0.3cm}
\end{table}
    
    To investigate whether our method possesses the scalability to indoor scene reconstructions, we \dan{further} conduct the experiments of 3D reconstruction from un-oriented point clouds on the synthetic indoor scene dataset \cite{peng2020convolutional}. 
    From the qualitative \dan{comparison shown} in Figure~\ref{fig:synroom}, we can \dan{observe} that some subtle legs of chairs can also be recovered by our approach, while others cannot capture these details. This demonstrates that ours can scale well to large scenes as we adopt the strategy of local geometry reasoning, instead of global shape modeling \dan{as} in SAL~\cite{atzmon2020sal} and IGR~\cite{gropp2020implicit}. 
    Compared to CONet~\cite{peng2020convolutional}, more fine-grained surface recoveries demonstrate that our approach can achieve better generality to novel scenes, due to the effective sign-agnostic optimization that conforms the desired implicit surface to the observed un-oriented surfaces.
    Moreover, the bypassing of normal estimation enables more robust scene-level surface reconstructions.
    Again, better quantitative results presented in Table~\ref{table:synthetic_room} consistently demonstrate the superiority of our approach. 
    
    
    \begin{table}[htbp]
    \centering
        \begin{tabular}{l|ccccc}
        \toprule
        Methods  & CD $\downarrow$ & NC $\uparrow $ & FS ($\tau$) $\uparrow$ & FS (2$\tau$) $\uparrow$  \\
        \hline\hline
        SPSR \cite{kazhdan2013screened} & 1.339 & 84.60  &  \textbf{82.33} & 87.83 \\
        \hline
        SAL \cite{atzmon2020sal} & 2.026 & 81.24  & 61.54 & 80.90  \\
        IGR \cite{gropp2020implicit}   & 2.392 & 84.12  &  78.07  & 83.98  \\
        \hline
        CONet \cite{peng2020convolutional}  & 1.559  & 82.05  & 59.55 & 80.76 \\
        LIG \cite{jiang2020local}      & 1.501  & 81.99   & 70.39 & 78.30  \\
        \hline
        Ours & \textbf{0.728} & \textbf{86.40} & 82.08  & \textbf{95.86}\\
        \bottomrule
        \end{tabular}
    \vspace{0.1cm}
    \caption{
    Quantitative \dan{comparison} for surface reconstruction from un-oriented point clouds on the real-world ScanNet dataset \cite{peng2020convolutional}. As watertight meshes are not provided by ScanNet, we directly evaluate all methods using the pretrained models on the synthetic room dataset. } 
    \label{table:scannet}
    \vspace{-0.3cm}
\end{table}

\begin{figure*}[t]

    \begin{tabular}{cc}
      \rotatebox[origin=c]{90}{
        \begin{tabular}{cccccc}
        \quad GT \quad \quad \quad \quad \quad \quad \quad &
            Ours \quad \quad \quad \quad \quad \quad \quad & 
            LIG\cite{jiang2020local} \quad \quad \quad \quad \quad \quad& 
            CONet\cite{peng2020convolutional} \quad \quad \quad \quad \quad \quad& 
            SPSR\cite{kazhdan2013screened} \quad \quad \quad \quad \quad & 
            Input PC 
        \end{tabular}
      } & 
      \hspace{-0.5cm}
      \raisebox{-.5\height}{\includegraphics[scale=0.165]{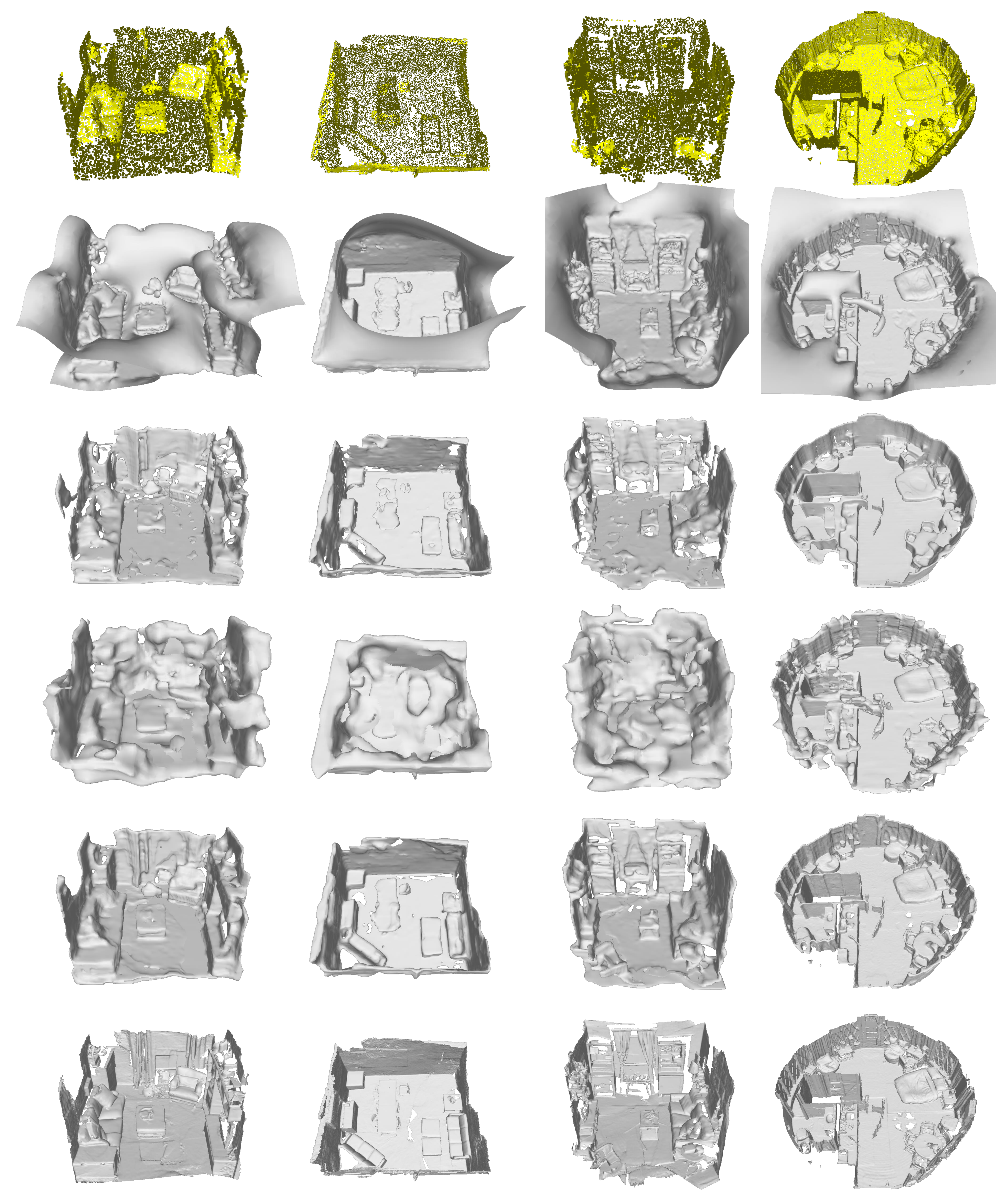}}
    \end{tabular}

\begin{tabular}{p{120pt}p{120pt}p{120pt}p{120pt}}
\hspace{3.0cm} (a) & \hspace{2.5cm} (b) & \hspace{1.8cm} (c) & \hspace{0.9cm} (d)
\end{tabular}

    \caption{\textbf{Scene-level Reconstruction on ScanNet~\cite{dai2017scannet} and Matterport3D~\cite{chang2017matterport3d}.} Qualitative \dan{comparison} for surface reconstruction from un-orientated scans of ScanNet (a, b, c) and Matterport3D (d). All methods except SPSR are trained on the synthetic room dataset and directly evaluated on ScanNet.}
    \label{fig:scannet}
\end{figure*}

\section{Real-World Scenes Generalization}
To compare the generalization performance on real-world scans, we also evaluate our approach on the real-world datasets, including ScanNet~\cite{dai2017scannet} and Matterport3D~\cite{chang2017matterport3d}. Notably, all models are only trained on the \dan{same} synthetic indoor scene dataset. 

\noindent \textbf{ScanNet-V2} The qualitative and quantitative comparisons are respectively shown in Figure \ref{fig:scannet} and Table \ref{table:scannet}. As \dan{can be} seen, compared to other methods, our results achieve \dan{clearly} better numerical scores and \dan{more fine-grained} surface geometries, which \dan{effectively verifies} the better generalization ability of \dan{the} proposed method on real-world scans.

\noindent \textbf{Matterport3D} To evaluate its scalability to huge scenes that contain multiple rooms, we finally conduct experiments on the Matterport3D dataset~\cite{chang2017matterport3d}. Following the sliding-window strategy presented in~\cite{peng2020convolutional}, we separately apply the designed sign-agnostic optimization of convolutional occupancy networks to each room. The visualization \dan{comparison is} presented in Figure~\ref{fig:scannet} (d).
Notably, the Matterport3D~\cite{chang2017matterport3d} is significantly different from the synthetic indoor room dataset that is used to pre-train our network. But our reconstruction results can still preserve rich details inside each room while adhering to the room layout, which \dan{fully} demonstrates that our method can achieve better scalability to huge scenes and better robustness to noises from \dan{different} sensing devices.


\section{Conclusion}
 For the task of surface reconstruction from un-oriented point clouds, we have proposed a simple yet effective solution of learning implicit surface reconstructions by sign-agnostic optimization of convolutional occupancy networks, \tang{which achieves scalability to large scenes, generality to novel shapes, and applicability to real-world scans in a unified framework. The characteristics of implicit field learning from convolutional features of hourglass networks} enable the test-time optimization without the use of surface normals.
 Extensive experiments on both object-level and scene-level datasets show that our method significantly outperforms the existing methods, both quantitatively and qualitatively.
 A limitation of our approach is the slow inference speed, which is also a common drawback of test-time optimization methods. We leave it as our future effort.

\par\noindent\textbf{Acknowledgement.} 
This work was partially supported by the Guangdong R$\&$D key project of China (No.: 2019B010155001), the National Natural Science Foundation of China (No.: 61771201), the Program for Guangdong Introducing Innovative and Entrepreneurial Teams (No.: 2017ZT07X183), the Early Career Scheme of the Research Grants Council (RGC) of the Hong Kong SAR GRF (No.: 26202321), HKUST Startup Fund (No.: R9253), and Alibaba DAMO Academy.


\begin{appendix}
\section*{Appendix}
In this supplementary material, we provide more details about our network \dan{architecture} in Section~\ref{SecNet}. Then we present ablation studies to validate the effectiveness of each design in our approach in Section~\ref{SecAbla}. \dan{In the next}, we demonstrate the generalization capabilities of our approach to novel categories that are different from the training category (``chair'') in Section~\ref{SecGen}. Finally, we show more qualitative \dan{comparison} with other competitive methods on the real-world \dan{3D} scene datasets in Section~\ref{SecReal}.

\section{Network Architectures}
\label{SecNet}
        \begin{figure*}[t]
        \centering
        \includegraphics[scale=0.46]{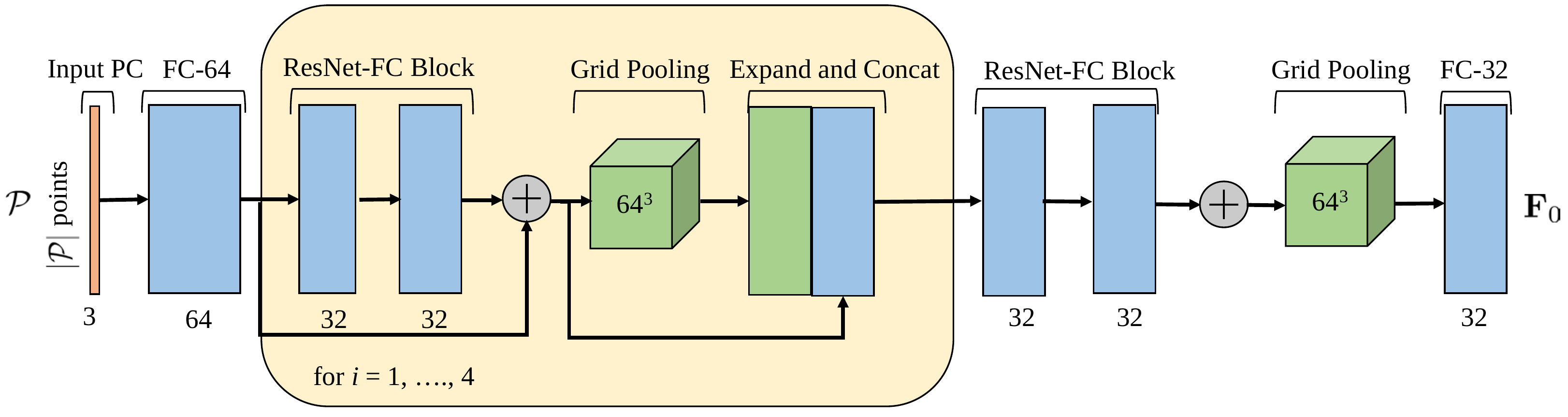}
        \caption{\textbf{ResNet~\cite{he2016deep} variants of PointNet\cite{charles2017pointnet}}. It utilizes a stack of five ResNet-FC blocks with skip connections and grid-pooling layers to extract point-wise features $\mathbf{F_0}$ from the observed surface point cloud $\mathcal{P}$.}
    \label{fig:pointnet}
    \end{figure*}
    
    \begin{figure*}[t]
    \vspace{2pt}
        \centering
        \includegraphics[scale=0.48]{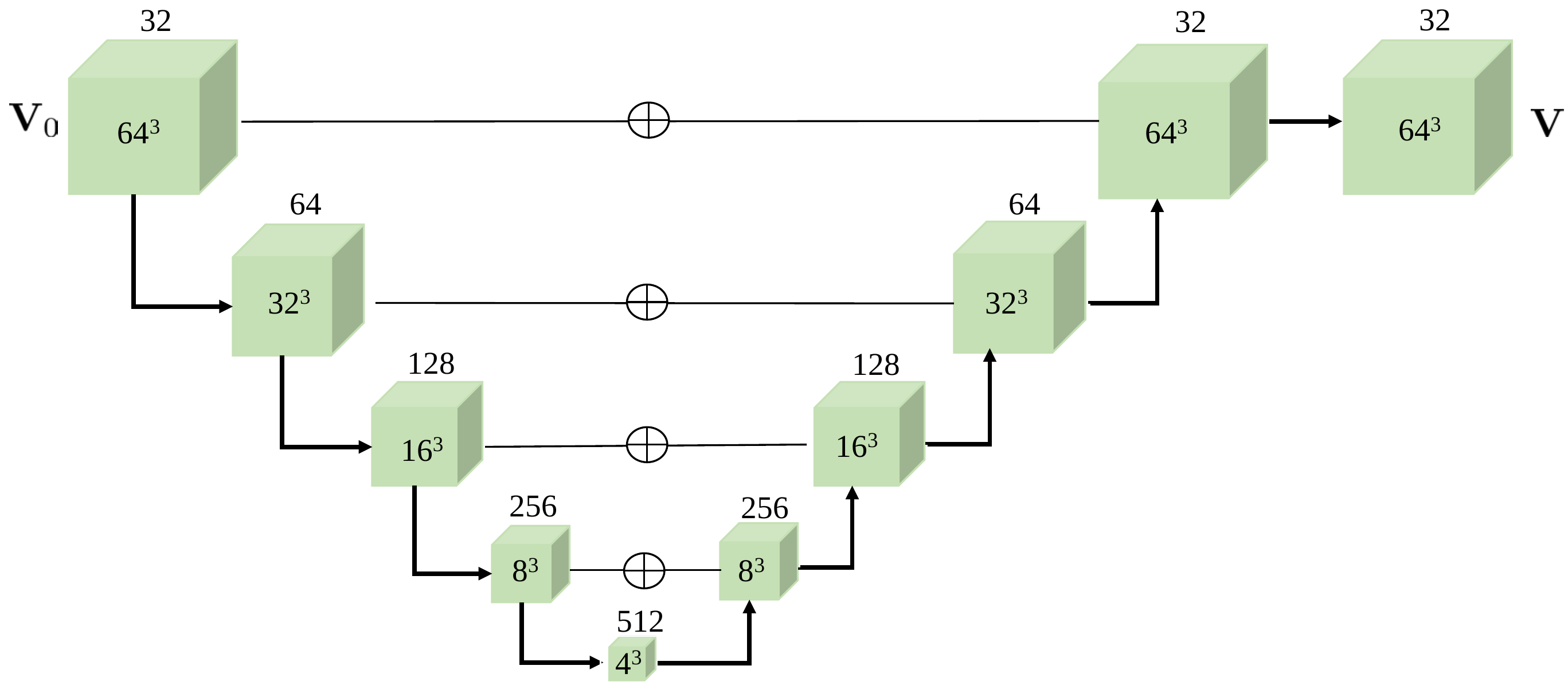}
        \caption{\textbf{3D U-Net.} To effectively fuse the global and local information of input shape, we transform $\mathbf{V}_0$ (produced from $\mathbf{F}_0$) to $\mathbf{V}$ using a 3D U-Net, which consists of a series of 3D down- and up-sampling convolutions with skip connections.}
    \label{fig:unet}
    \vspace{2pt}
    \end{figure*}
    
    \begin{figure}[h]
        \centering
        \includegraphics[scale=0.52]{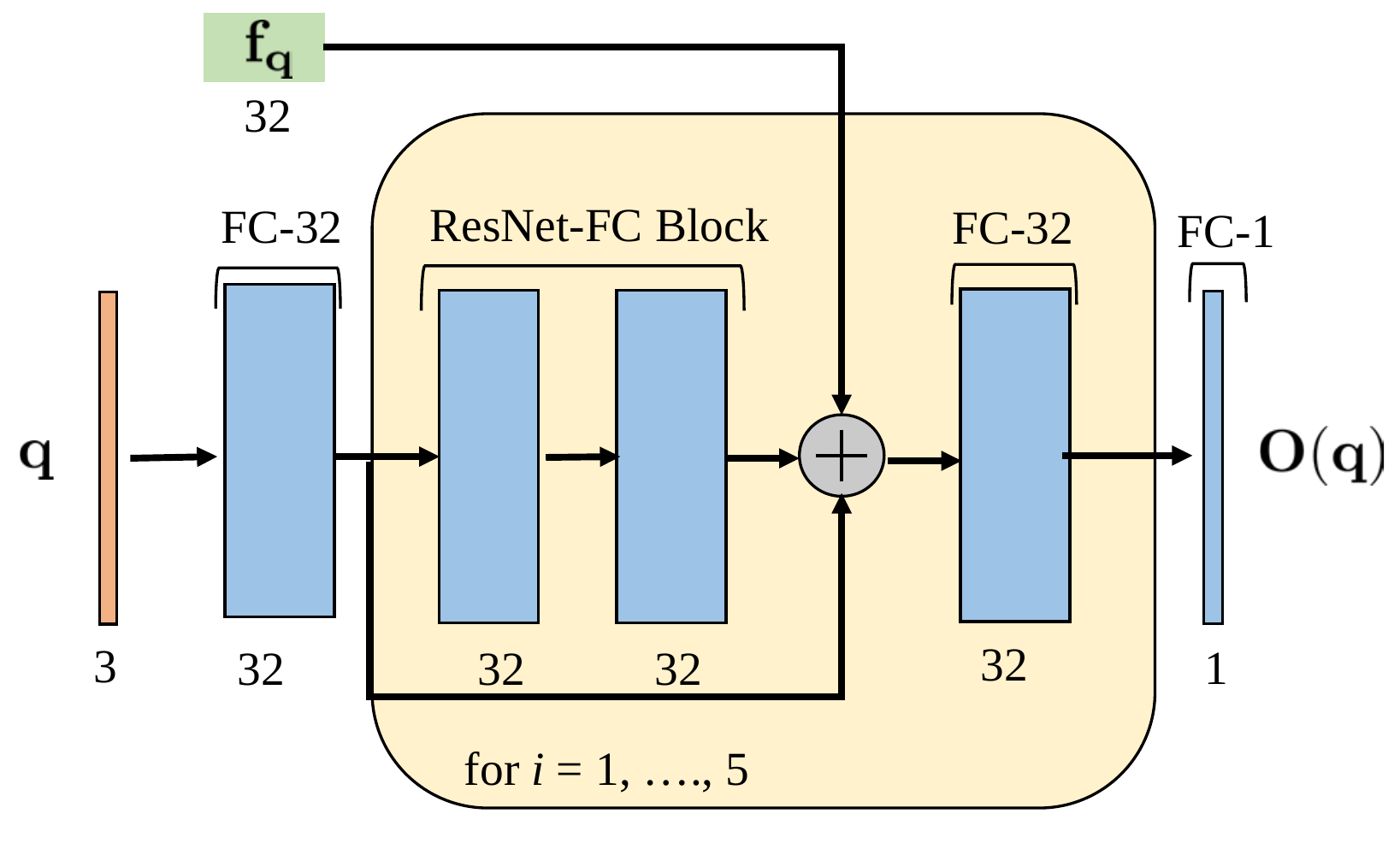}
        \caption{\textbf{Occupancy Decoder.} It contains five ResNet-FC blocks with skip connections. Given a point $\mathbf{q}$ randomly sampled in the 3D space, we query a feature vector $\mathbf{f_q}$ from the feature volume $\mathbf{V}$ according to the location of $\mathbf{q}$. Then we pass $\mathbf{q}$ and $\mathbf{f_q}$ into the occupancy decoder to predict the occupancy probability of $\mathbf{q}$ (\ie $\mathbf{O_q}$).}
    \label{fig:decoder}
    \end{figure}

    \noindent \textbf{PointNet}: The detailed network architecture of PointNet used in the paper is depicted in Figure~\ref{fig:pointnet}. Firstly, we map the coordinates of $\mathcal{P}$ into the feature space using a fully-connected (FC) layer and a ResNet-FC~\cite{he2016deep} block. Then, instead of using a global pooling operation to obtain a global feature like~\cite{charles2017pointnet}, we perform the grid-pooling operation~\cite{liao2018deep} to locally fuse the extracted features. Specifically, we perform an average-pooling operation for the features that are within the same voxel cell from a volumetric grid with the size of $64^3$. Next, we concatenate the locally pooled features with the features before pooling, and then feed the formed features into the subsequent ResNet-FC block. Overall, we use 5 ResNet blocks with intermediate grid-pooling layers to obtain the point-wise features $\mathbf{F}_0$.
    
    \noindent \textbf{3D U-Net}: The network architecture of 3D-UNet is illustrated in Figure~\ref{fig:unet}. The 3D U-Net~\cite{ronneberger2015u} is used to aggregate both local and global information of the volumetric feature $\mathbf{V}_0$ that is transformed from $\mathbf{F}_0$. The dimensions of input and output features are both set to 64. To ensure that the receptive field is equal to or larger than the size of the input feature volume, the depth of the 3D U-Net is set to 4. 
    
    \noindent \textbf{Occupancy Decoder}: As shown in Figure~\ref{fig:decoder}, the occupancy decoder consists of 5 stacked ResNet-FC blocks with skip connections. And the hidden feature dimension is set to 32.

\section{Ablation studies}
\label{SecAbla}
    
    \begin{figure}[t]
        \vspace{5pt}
        \centering
        \includegraphics[scale=0.51]{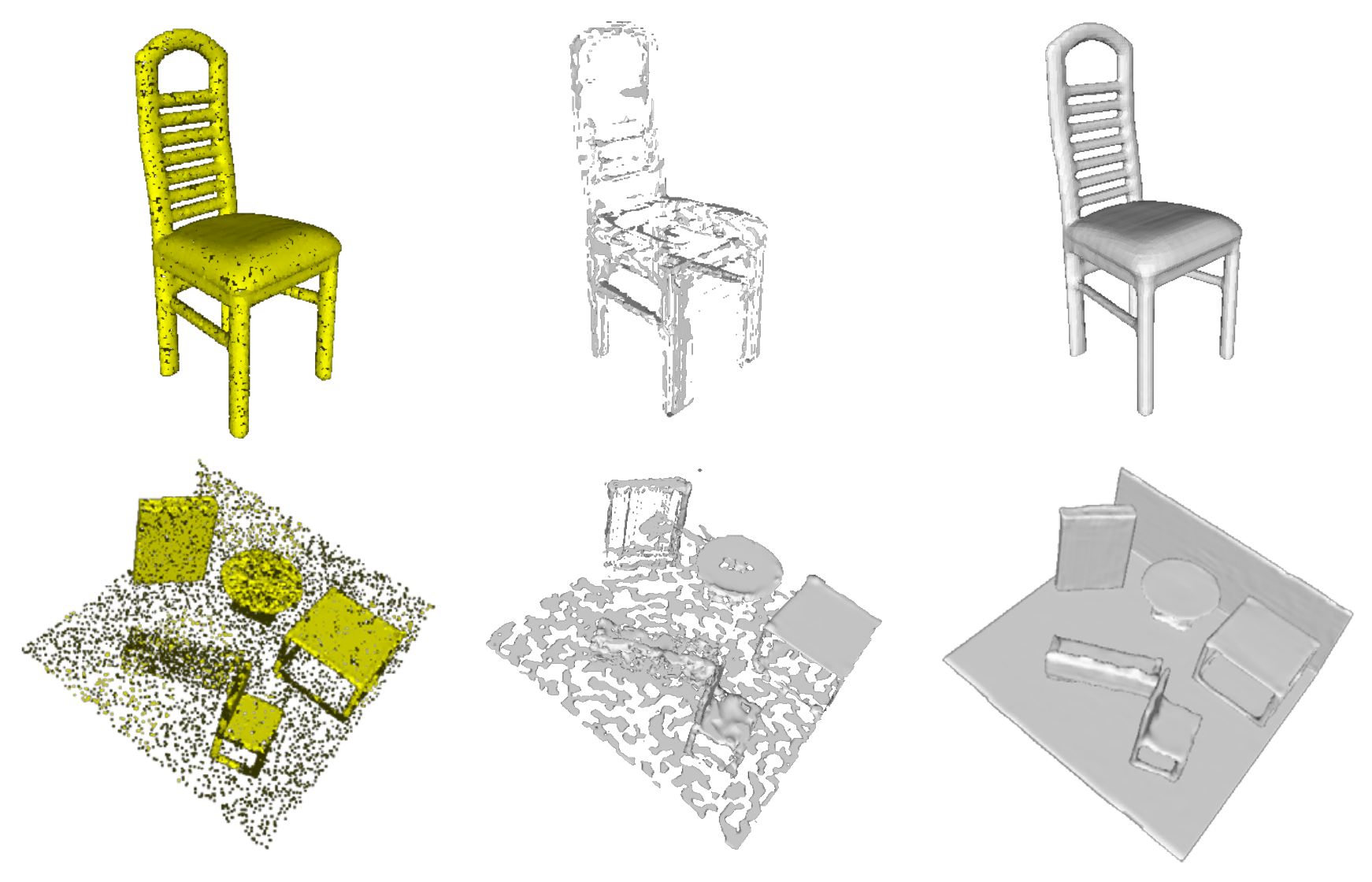}
        
        \begin{tabular}{p{80pt}p{80pt}p{80pt}}
            \hspace{0.3cm} (a) & \hspace{0.4cm} (b) &\hspace{0.4cm} (c)
        \end{tabular}
        \caption{\textbf{Additional Qualitative Ablation Studies:} (a) input point clouds, (b) without the pre-training of convolutional occupancy networks, and (c) Ours.}
    \label{fig:ablation}
    \end{figure}
    In this section, we conduct additional ablation studies by alternatively removing one of the modules of the proposed approach to verify the effectiveness of them.
    
\noindent \textbf{Effect of pre-training (\ie~w/o pre-training)}
    Based on our approach, an alternative solution to provide \dan{initialization of the} signed field for the proposed sign-agnostic optimization is to adopt the geometric initialization as in SAL~\cite{atzmon2020sal}, which initializes the implicit decoder to approximate the signed distance field of the unit sphere. The visualization comparisons are shown in Figure~\ref{fig:ablation}. Without the pre-trained shape prior, the sign agnostic optimization fails to reconstruct reasonable geometries.

    \noindent \textbf{Effect of only optimizing the encoder (\ie~opt. enc.)} In all experiments, we choose to optimize the whole network parameters with the unsigned binary cross-entropy loss during inference. An alternative solution is to only optimize the encoder (\ie~PointNet and 3D U-Net) while freezing the occupancy decoder. The comparisons shown in Table~\ref{table:ablation}  clearly demonstrate that jointly optimizing the whole network can achieve better generality to unseen shapes.
    
    \begin{table}[h]
    \renewcommand\arraystretch{1.2}
    \setlength{\tabcolsep}{1.5mm}
        \centering
        \scalebox{0.9}{
            \begin{tabular}{l | l |c c c c}
            \toprule
            Datasets & Methods  & CD $\downarrow$ & NC $\uparrow $ & FS ($\tau$) $\uparrow$ &  FS (2$\tau$) $\uparrow$ \\ 
            \hline \hline
             \multirow{2}{*}{\shortstack{ShapaNet \\ -chair~\cite{chang2015shapenet}}}         & opt. enc. &  \textbf{0.516}  &  93.42  &   97.15 &  \textbf{99.40} \\
              & Ours & 0.522 & \textbf{93.51} & \textbf{97.16} & 99.37 \\ 
            \hline
            \multirow{2}{*}{\shortstack{Synethetic \\ Room~\cite{peng2020convolutional}}} & opt. enc. & 0.516  &  89.75  &  93.43  &  98.53 \\
                  & Ours & \textbf{0.495} & \textbf{90.04} &  \textbf{93.85} & \textbf{98.82}   \\
            \hline
            \multirow{2}{*}{ScanNet~\cite{dai2017scannet}}          & opt. enc. & 0.741  & 86.24 & 81.49 & 95.56 \\
               &Ours & \textbf{0.728} & \textbf{86.40} & \textbf{82.08}  & \textbf{95.86}\\
            \bottomrule
            \end{tabular}
        }
        \caption{\textbf{Additional ablation studies} on three datasets.} 
        \label{table:ablation}
        \vspace{-5pt}
\end{table}
    
\begin{figure}[h]
    	\centering
    	\includegraphics[width=1.0\linewidth]{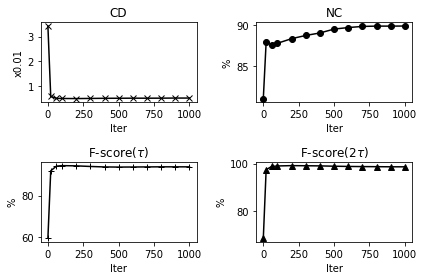}
    	\caption{Quantitative results obtained at different iterations during the test-time optimization. {Experiments are conducted on the synthetic room dataset with the input of 30,000 points.}}
    	\label{fig:vary_iter:quanti}
    \end{figure}

\begin{figure}[h]
    	\centering
    	\includegraphics[width=1.0\linewidth]{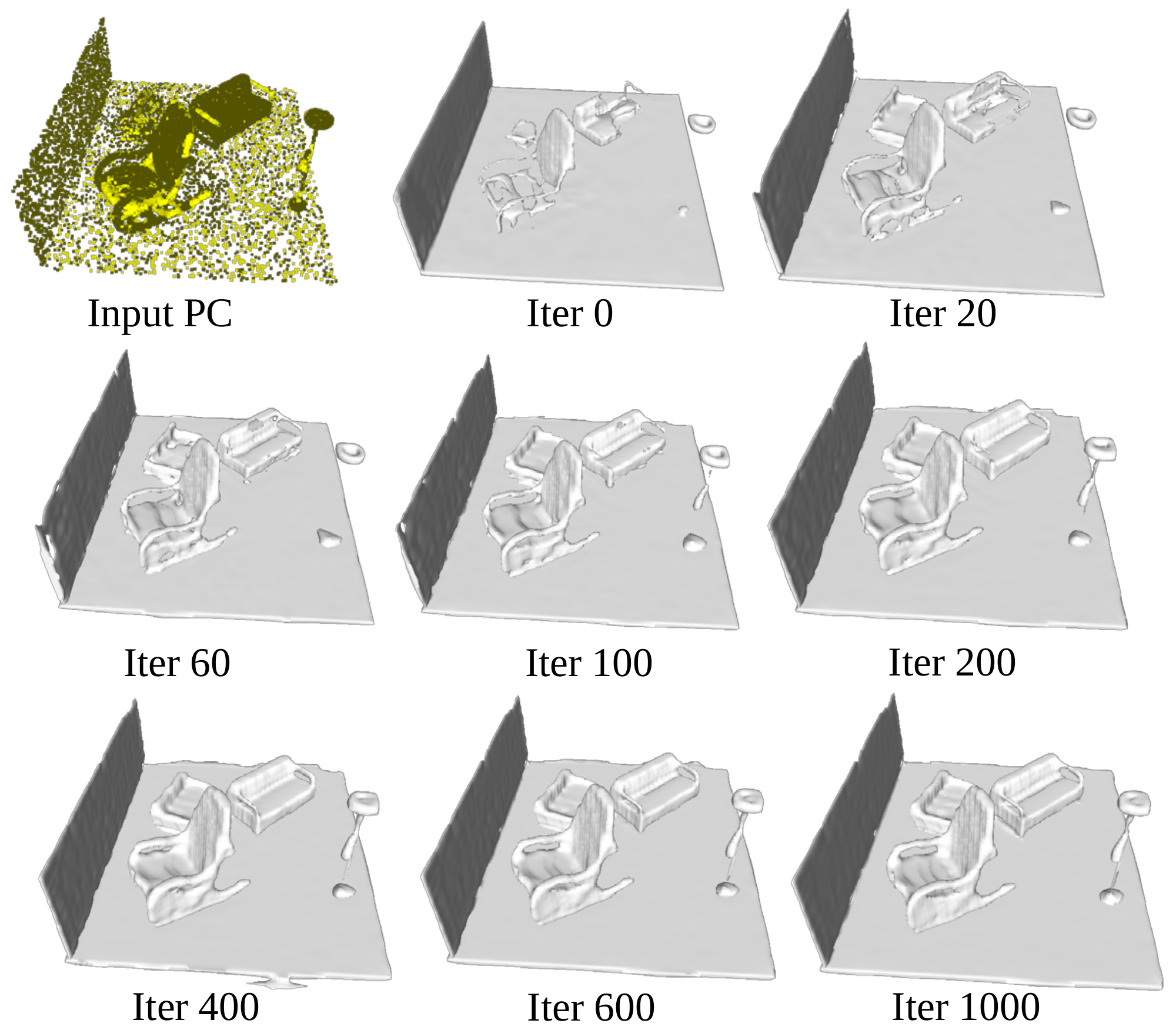}
    	\caption{Examples of qualitative results of {a synthetic room} obtained at different iterations during the test-time optimization.}
    	\label{fig:vary_iter:quali}
    \end{figure}

    \noindent\textbf{Ablation studies on the iteration number of the test-time optimization.} Fig.~\ref{fig:vary_iter:quanti} and~\ref{fig:vary_iter:quali} show the quantitative and qualitative results w.r.t. the number of iterations, respectively. {Notably, the `Iter 0' represents the result before optimization.} We can observe that after about 600 iterations of the test-time optimization, the results become stable.
    
    \noindent\textbf{Ablation studies on the sparsity level of the input.} Quantitative results for different sparsity levels of the input are shown in Table~\ref{table:sparsity:chair} and ~\ref{table:sparsity:synthetic_room}. We can observe that the results of different evaluation metrics only show a slightly small variance, which clearly demonstrates the robustness of our method against the input sparsity.
    
        \begin{table}[htbp]
        \centering
        \resizebox{0.83\linewidth}{!}{
        \begin{tabular}{l|cccc}
            \toprule
             $|\mathcal{P}|$  & CD $\downarrow$ & NC $\uparrow $ & FS ($\tau$) $\uparrow$ & FS (2$\tau$) $\uparrow$  \\
            \hline\hline
            5,000   & 0.529 & 89.71 & 95.51 & 99.00   \\
            10,000  & 0.524 & 92.37 & 96.85 & 99.20   \\
            20,000  & 0.522 & 93.29 & 97.11 & 99.06   \\
            30,000  & 0.522 & 93.51 & 97.16 & 99.37   \\
            40,000  & 0.502 & 93.57 & 97.11 & 99.35   \\
            50,000  & 0.502 & 93.61 & 97.04 & 99.29   \\
            \bottomrule
        \end{tabular}
        }
        \vspace{1mm}
        \caption{ Quantitative results at different sparsity levels of
        the input point cloud $\mathcal{P}$ on the ShapeNet `chair' category.}
        \vspace{-0.2cm}
        \label{table:sparsity:chair}
    \end{table}

    \begin{table}[htbp]
        \centering
        \resizebox{0.83\linewidth}{!}{
        \begin{tabular}{l|cccc}
            \toprule
             $|\mathcal{P}|$  & CD $\downarrow$ & NC $\uparrow $ & FS ($\tau$) $\uparrow$ & FS (2$\tau$) $\uparrow$  \\
            \hline\hline
            5,000   & 0.511 & 89.24  & 93.50 &  98.57   \\
            10,000  & 0.494 & 89.86  & 94.01 &  98.89   \\
            20,000  & 0.494 & 90.03  & 93.85 &  98.73   \\
            30,000  & 0.495 & 90.04  & 93.85 &  98.82   \\
            40,000  & 0.488 & 90.04  & 93.85 &  98.73   \\
            50,000  & 0.476 & 89.98  & 93.95 &  98.99   \\
            \bottomrule
        \end{tabular}
        }
        \caption{ Quantitative results at different sparsity levels of
        the input point cloud $\mathcal{P}$ on the synthetic room dataset.}
        \label{table:sparsity:synthetic_room}
    \end{table}

\section{Novel Categories Generalization}
\label{SecGen}
    \begin{figure*}[t]
    \vspace{5pt}
        \centering
        \includegraphics[scale=0.128]{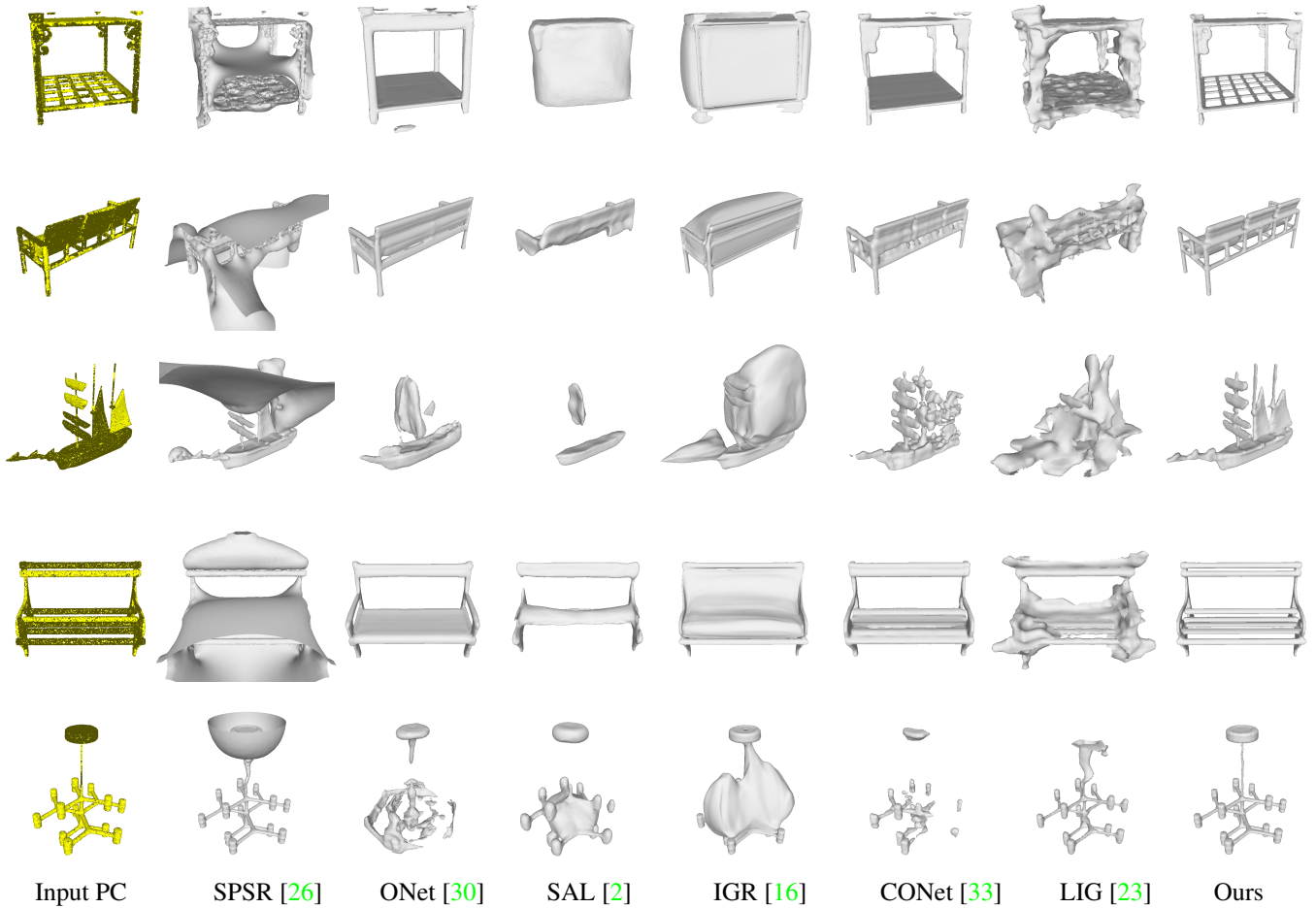}
        
        \begin{tabular}{p{60pt}p{53pt}p{53pt}p{53pt}p{53pt}p{53pt}p{53pt}p{53pt}}
                 \ \ Input PC  & 
                 \  SPSR \cite{kazhdan2013screened} &
                 \ ONet \cite{mescheder2019occupancy} & 
                 \  SAL \cite{atzmon2020sal}  & 
                 \ IGR \cite{gropp2020implicit} & 
                 \ CONet \cite{peng2020convolutional}  &
                 \ \ \ LIG \cite{jiang2020local} &  
                 \ Ours
        \end{tabular}
        \caption{\textbf{Generalize to Novel Categories.} We directly evaluate our approach and baselines on unseen, novel categories including ``bench'', ``lamp'', and ``watercraft'' that are very different from the training ``chair'' category.}
    \label{fig:novelcat}
    \end{figure*}
    In this section, we analyze the generalization performance of our approach and the baselines on the object-level reconstruction. We directly evaluate them on novel categories such as ``bench'', ``lamp'' and ``watercraft'' that are different from the training ``chair'' category. As shown in Figure~\ref{fig:novelcat}, our approach can preserve more geometric details such as small holes, long rods, and thin parts, while the baselines cannot. This demonstrates the superior generalization capabilities of the proposed approach to unseen categories.

\section{Real-world Scenes Generalization}
\label{SecReal}
    In this section, we \dan{first} describe the implementation details of sign-agnostic optimization of convolutional occupancy networks in a sliding-window manner, and then provide more qualitative \dan{comparison} on the real-world scenes \dan{datasets including} ScanNet-V2~\cite{dai2017scannet} and Matterport3d~\cite{chang2017matterport3d}.
    
    \subsection{Implementation Details of Sign-Agnostic Optimization in a Sliding-Window Manner}
    In the experiments of object-level and synthetic scene reconstruction, we perform pre-training and sign-agnostic optimization within the unit cube. However, this strategy cannot deal with real-world scenes that are arbitrarily sized and represented in meters.
    Although we can resize these scenes into the unit cube, convert them into volumetric grids of size $64^3$, and then process them using the 3D U-Net as described in Section~\ref{SecNet}, we may not be able to recover fine-grained geometries as the low-resolution voxelization process loses much information about surface details, while the high-resolution voxelization such as $128^3, 256^3$ would suffer from the heavy computation cost and memory issues.
    Thanks to the \dan{translation equivalence} of fully convolutional networks, we can apply the proposed model to local patches cropped from large scenes and perform implicit surface reconstruction in a sliding-window manner, which can help us preserve the input information while avoiding memory issues of 3D CNNs.
    
    More specifically, we also pre-train our model on the synthetic indoor scene dataset~\cite{peng2020convolutional} where the size of scenes is approximately a real-world unit of 4.4m $\times$ 4.4m $\times$ 4.4m. Similar to the setting of~\cite{peng2020convolutional}, we set the voxel size as 0.02m such that each scene is contained in a volumetric grid with size $220^3$. 
    During the network pre-training, we utilize the Res-PointNet and 3D U-Net described in Section~\ref{SecNet} to learn corresponding convolutional features from each cropped subvolume. Then we predict the occupancy probabilities of query points uniformly sampled from the grid of input subvolume. 
    Specifically, we randomly sample one point within the whole scene and use it as the center of the subvolume. 
    The size of each cropped subvolume~(\ie $H \times W \times D$) is set to $25 \times 25 \times 25$. 
    Since the receptive field of 3D U-Net is 64, we set the size of input subvolumes to $(H + 63) × (W + 63) + (D + 63) = 88 \times 88 \times 88$. At each iteration, we use a batch size of 4 subvolumes.
    
    During the test-optimization stage, we divide the large scene into overlapped subvolumes and then perform sign-agnostic optimization for each subvolume in a sliding-window manner. We determine the size of cropped subvolumes according to the size of input scenes such that they are compatible with the GPU memory. Notably, we do not need the padding operation as the cropped subvolumes overlap.
    
    \subsection{Additional Qualitative Results}
\begin{figure*}[t]
        \begin{tabular}{cc}
          \rotatebox[origin=c]{90}{
            \begin{tabular}{c c c c c c}
            \quad \quad GT \quad \quad \quad \quad \quad \quad  \quad &
                Ours \quad \quad \quad \quad \quad \quad  & 
                LIG\cite{jiang2020local} \quad \quad \quad \quad \quad  \quad \quad &
                CONet\cite{peng2020convolutional} \quad \quad \quad  \quad \quad  & 
                SPSR\cite{kazhdan2013screened} \quad \quad \quad \quad \quad & 
                Input PC \quad \quad \quad 
            \end{tabular}
          } & 
          \hspace{-0.5cm}
          \raisebox{-.5\height}{\includegraphics[scale=0.165]{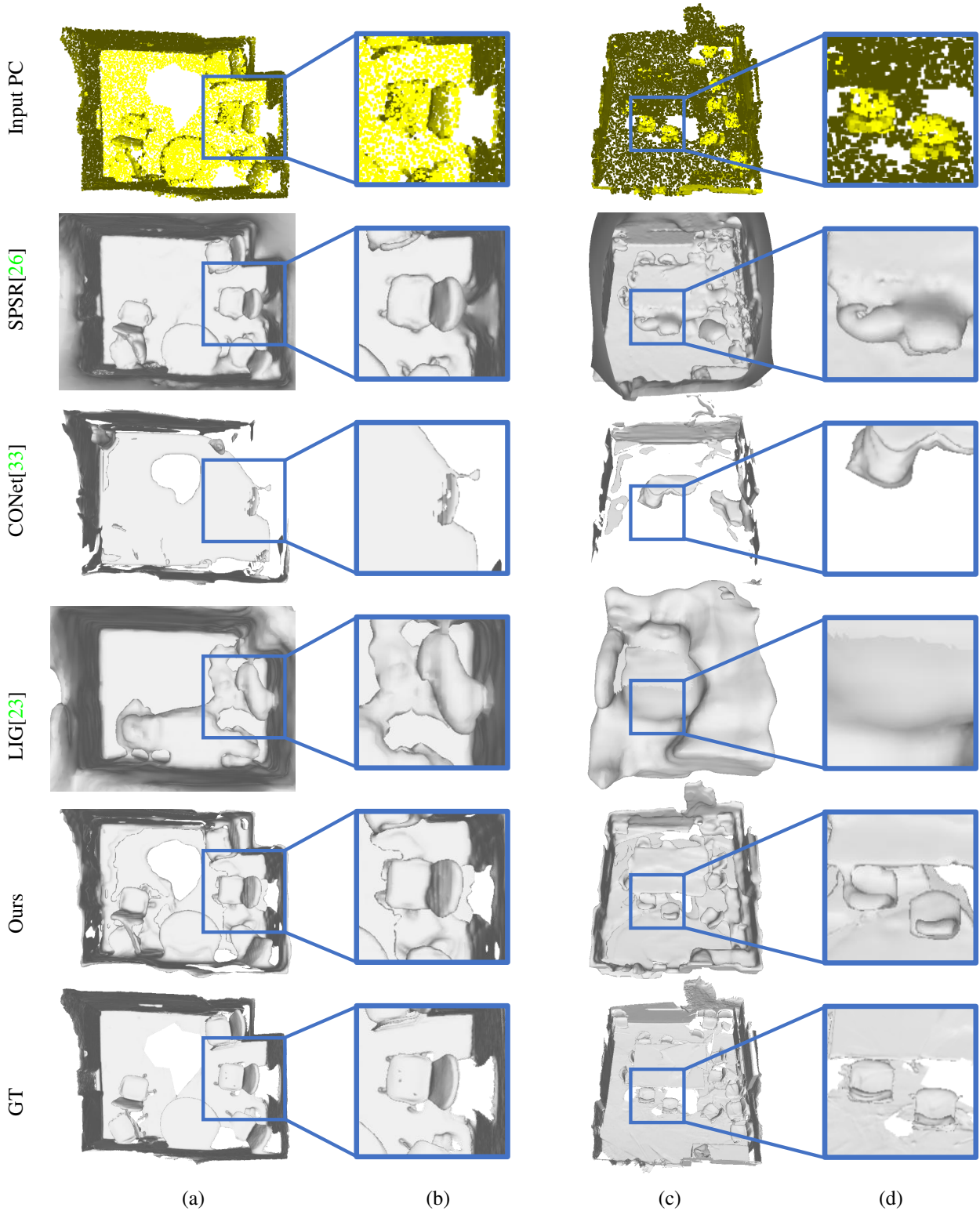}}
        \end{tabular}
    
    \begin{tabular}{p{120pt}p{120pt}p{120pt}p{120pt}}
    \hspace{3.0cm} (a) & \hspace{2.5cm} (b) & \hspace{1.8cm} (c) & \hspace{0.9cm} (d)
    \end{tabular}
        \caption{\textbf{Scene-level Reconstruction on ScanNet~\cite{dai2017scannet}.} Qualitative comparisons for surface reconstruction from un-orientated scans of ScanNet. All methods except SPSR are trained on the synthetic room dataset and directly evaluated on ScanNet.}
        \label{fig:scannet_fig2}
    \end{figure*}

\begin{figure*}[t]
        \begin{tabular}{cc}
          \rotatebox[origin=c]{90}{
            \begin{tabular}{c c c c c c}
            \quad GT \quad \quad \quad \quad \quad \quad \quad &
                Ours \quad \quad \quad \quad \quad \quad \quad & 
                LIG\cite{jiang2020local} \quad \quad \quad \quad \quad \quad& 
                CONet\cite{peng2020convolutional} \quad \quad \quad \quad \quad \quad& 
                SPSR\cite{kazhdan2013screened} \quad \quad \quad \quad \quad & 
                Input PC 
            \end{tabular}
          } & 
          \hspace{-0.5cm}
          \raisebox{-.5\height}{\includegraphics[scale=0.165]{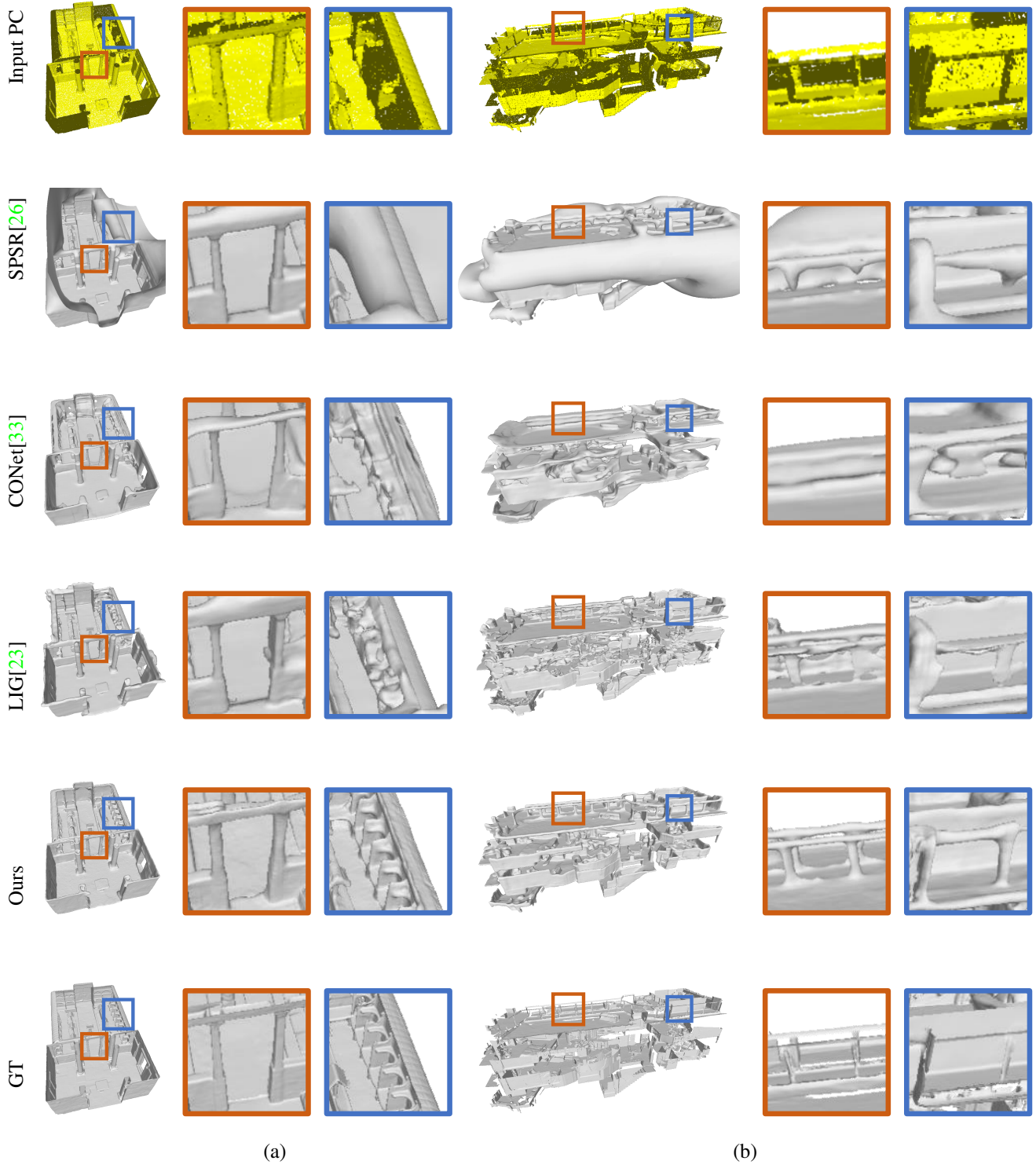}}
        \end{tabular}
    
    \begin{tabular}{p{200pt}p{200pt}}
    \hspace{3.0cm} (a) & \hspace{3.5cm} (b) 
    \end{tabular}
        \caption{\textbf{Scene-level Reconstruction on Matterport 3D~\cite{chang2017matterport3d}.} Qualitative comparisons for surface reconstruction from un-orientated scans of  Matterport3D. All methods except SPSR are trained on the synthetic room dataset and directly evaluated on Matterport 3D.}
        \label{fig:matterport_fig1}
    \end{figure*}

    We have provided more qualitative comparisons on the ScanNet~\cite{dai2017scannet} in Figure~\ref{fig:scannet_fig2}.
    Besides, more visualized results on the Matterport3D~\cite{chang2017matterport3d} are shown in Figure~\ref{fig:matterport_fig1}. 
    From these results, we can clearly observe that our method achieves more superior performance to large scenes with multiple rooms than the existing state-of-the-arts.
    And in comparison with those baselines such as SPSR~\cite{kazhdan2013screened, jiang2020local} that heavily rely on accurate surface normals, our approach can avoid the degenerated results caused by inaccurate normal estimation. Besides, compared to CONet~\cite{peng2020convolutional}, our approach can reconstruct more complete geometries and preserve complicated geometric details well, which validates the effectiveness of the proposed sign-agnostic optimization during inference. Overall, our method simultaneously maximizes the scalability to large scenes, generality to unseen shapes, and applicability to real scans that lack reliable surface normals.

    
    
\end{appendix}
\clearpage
{\small
\bibliographystyle{ieee_fullname}
\bibliography{egbib}
}

\end{document}